# Benchmarking Distilled Language Models: Performance and Efficiency in Resource-Constrained Settings

Sachin Gopal Wani, Eric Page, Ajay Dholakia, and David Ellison

Lenovo, Infrastructure Solutions Group, Morrisville, NC, USA
{swani1, epage2, adholakia, dellison}@lenovo.com

**Abstract.** Knowledge distillation offers a transformative pathway to developing powerful, yet efficient, small language models (SLMs) suitable for resource-constrained environments. In this paper, we benchmark the performance and computational cost of distilled models against their vanilla and proprietary counterparts, providing a quantitative analysis of their efficiency.

Our results demonstrate that distillation creates a superior performance-to-compute curve. We find that creating a distilled 8B model is over 2,000 times more compute-efficient than training its vanilla counterpart, while achieving reasoning capabilities on par with, or even exceeding, standard models ten times its size. These findings validate distillation not just as a compression technique, but as a primary strategy for building state-of-the-art, accessible AI.

**Keywords:** Distillation, Small Language Models, Resource-Limited Settings, Computational Efficiency, Benchmarking, Model Evaluation, Large Language Models, Generative AI.

## 1 Introduction

The advent of Large Language Models (LLMs) has marked a transformative era in artificial intelligence, demonstrating remarkable capabilities in natural language understanding, generation, and reasoning. The performance of these models has consistently scaled with their size, leading to the development of foundation models with hundreds of billions, or even trillions, of parameters. However, this relentless pursuit of scale comes at a staggering computational cost. The immense resources required for pre-training and fine-tuning are significant, but even deploying these massive models for inference demands substantial computational power. This creates significant barriers, limiting their use to a handful of well-resourced corporations and rendering them impractical for widespread adoption in real-world applications that require low latency and cost efficiency.

The challenge, therefore, is to create models that are not only compute-efficient but also possess the sophisticated capabilities required for specialized tasks. Standard training or fine-tuning approaches on smaller models often fail to replicate the nuanced reasoning and generalization abilities of their larger counterparts. This capability gap underscores the need for more advanced techniques that can efficiently imbue smaller models with the deep knowledge learned by state-of-the-art foundation models, paving the way for a new generation of powerful, yet accessible, AI systems.

Knowledge distillation has emerged as a powerful and promising technique to achieve this goal [1]. The fundamental principle involves transferring the knowledge from a large, powerful "teacher" model to a smaller, more efficient "student" model. By training the student on the rich, soft-target outputs generated by the teacher, it can learn nuanced patterns and capabilities that are difficult to acquire through standard training on a dataset alone. This process allows for the creation of compact models that retain a significant portion of their larger counterparts' performance while drastically reducing computational overhead for inference and training.

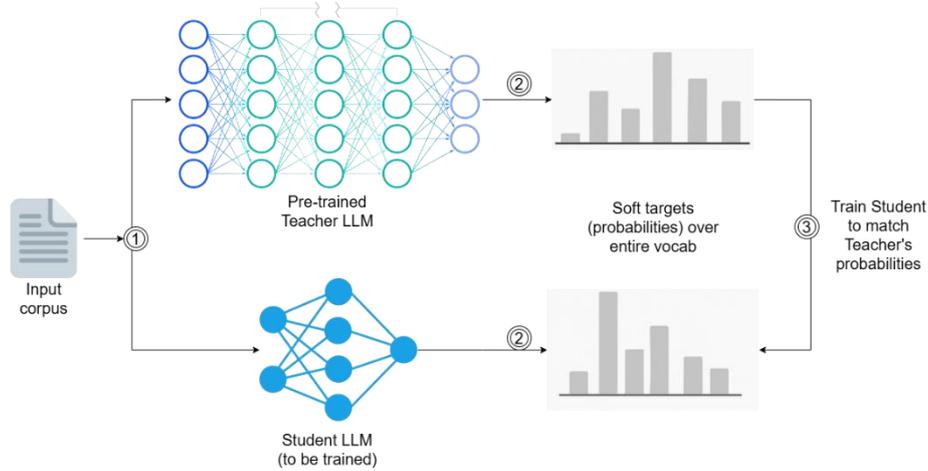

**Fig. 1.** Illustration of the Distillation process

In this paper, we conduct a comparative meta-benchmark to quantify the trade-offs between knowledge distillation and other common model adaptation techniques. Through a rigorous analysis of published benchmarks and computational models, we evaluate the costs—measured in Floating Point Operations (FLOPs) and GPU-hours—alongside performance on standard tasks. We ground our analysis in concrete examples, such as the distillation of the DeepSeek-R1 model [2], to provide a clear, evidence-based framework for evaluating model efficiency. Our findings demonstrate that knowledge distillation offers a highly compute-efficient pathway for developing powerful models, positioning it as a preferred strategy for building and deploying generative AI systems within the operational and financial constraints of real-world environments.

## 2    Related Work

### 2.1    Fine-Tuning

Fine-tuning a LLM involves taking a pre-trained model and further training it on a smaller, more specialized dataset to improve its performance in a specific domain, such as question answering. Classical fine-tuning involves training a portion or the entire model, but due to the increasing size of LLMs, Parameter-Efficient Fine-Tuning (PEFT) methods have been developed to reduce memory and compute costs [3, 9, 10]. However, while fine-tuning has been shown to adjust a model's style, it doesn't imbue the model with new knowledge, especially outside of the fine-tuning dataset's domain [11]. This inherent limitation of fine-tuning—its inability to instill fundamentally new reasoning pathways—highlights the need for alternative methods like knowledge distillation to more effectively transfer capabilities from superior models.

## 2.2 Knowledge Distillation

Knowledge Distillation involves transferring the capabilities of a large "teacher" model into a smaller "student" one by having the student mimic the soft output distribution of the teacher instead of hard labels [1]. While originally applied to deep neural networks, this concept has been adapted to LLMs [12,13]. In practice, distillation has been shown to be very effective as distilled models outperform their baseline counterparts across a variety of tasks. Most notably, distilling reasoning model outputs and Chain-of-Thought (CoT) into smaller models significantly improves performance on reasoning tasks [2,14]. This is because the student learns not just the correct answer, but also the nuanced, step-by-step reasoning process encoded in the teacher's soft probability distribution.

## 2.3 Estimating Computational Cost

A quantitative comparison of these adaptation techniques requires a standardized metric for computational cost. The most widely accepted metric is the number of total floating-point operations (FLOPs) required for training. A well-established formula for estimating the compute cost of training a standard transformer-based LLM is provided by Epoch AI [4]:

$$\text{Training}_{\text{FLOPs}} \approx 6 \times N \times D$$

Where N is the number of parameters in the model and D is the number of tokens in the training dataset. The factor of 6 is a widely used approximation that accounts for both the forward pass (roughly $2 \times N \times D$ FLOPs) and the backward pass (roughly $4 \times N \times D$ FLOPs).

For knowledge distillation, the computational cost can be decomposed into two distinct components: the cost of generating the training data using the teacher model and the cost of training the student model.

1. Teacher Inference Cost: This step only requires a forward pass of the teacher model over the distillation dataset.

$$\text{TeacherInference}_{\text{FLOPs}} \approx 2 \times N_{\text{teacher}} \times D_{\text{distill}}$$

2. **Student Training Cost:** This involves both forward and backward passes for the smaller student model.

$$\text{StudentTraining}_{\text{FLOPs}} \approx 6 \times N_{\text{student}} \times D_{\text{distill}}$$

The total cost of distillation is the sum of these two components. This decomposition is critical for understanding the trade-offs, as the cost is dominated by the size of the teacher model and the distillation dataset, while the student training itself is a relatively small portion of the overall compute. This framework provides the basis for our comparative analysis of compute efficiency. Furthermore, the seminal work on scaling laws by Hoffmann et al. [5] (the "Chinchilla" paper) provides an empirical foundation for these compute-performance trade-offs, demonstrating that for a fixed compute budget, an optimal balance between model size and the amount of training data is crucial. This supports the core premise of our work: that creating smaller, more data-efficient models through methods like distillation can be a superior strategy to simply scaling up model size.

## 3     Methodology

To conduct a robust comparative analysis, we established a systematic methodology for estimating and comparing the computational costs associated with distillation, fine-tuning, and retraining. Our approach is grounded in established computational formulas and accounts for real-world hardware performance characteristics to translate abstract FLOPs into tangible GPU-hours and $CO_2$ emissions.

### 3.1    Estimating Computational Cost in FLOPs

Our primary metric for computational cost is the total number of floating-point operations (FLOPs). We use the formulas outlined in Section 2.3 for our calculations. A key challenge in applying these formulas is determining the size of the training dataset (D) in tokens. Our approach varied depending on the model adaptation technique:

- **For Retraining and Fine-Tuning:** We relied on token counts reported directly in the source literature or associated with benchmark datasets. This includes large-scale pre-training datasets (e.g., 15.6 trillion tokens for Llama3 models) and smaller, instruction-tuning datasets (e.g., ~1 billion domain specific tokens). For models that do not report the training data, we directly use the provided compute FLOPs in supporting literature.
- **For Distillation:** In cases like the DeepSeek-R1 model, where the source paper reports the number of training samples rather than tokens, we estimated the total token count. The distillation data is composed of approximately 600,000 reasoning-focused samples featuring Chain-of-Thought trajectories and 200,000 non-reasoning samples. Given the token-intensive nature of CoT, we adopt a weighted average of 750 tokens per sample for our calculations. This estimate reflects the heavy skew towards complex reasoning data and is consistent with the token lengths of typical instruction-following datasets. This allows us to convert sample counts into the token counts required for our FLOPs calculations.

### 3.2  Converting FLOPs to GPU-Hours

While FLOPs provide a hardware-agnostic measure of computation, GPU-hours offer a more practical metric for real-world cost and time estimation. To bridge this gap, we use the concept of Model FLOPs Utilization (MFU), which measures the ratio of the actual achieved throughput of a training job against the theoretical peak performance of the underlying hardware. MFU accounts for real-world bottlenecks such as memory bandwidth, network latency, and software overhead, which prevent hardware from reaching its peak theoretical output. For instance, the recent training of Llama 3.1 reported achieving an MFU of 38-43% on clusters of NVIDIA H100 GPUs [15]. This provides a concrete, empirically-backed efficiency range for our estimations.

Our final formula for estimating GPU-hours is:

$$\text{GPU-Hours} = \frac{\text{TotalFLOPs}}{\text{PeakFLOPs} \times \text{MFU} \times 3600}$$

This sophisticated approach allows us to generate credible time and cost estimates for running different training jobs on specific, real-world hardware. We analyze several leading data center GPUs, including NVIDIA's A100, H100, H200, and L40S, using their peak theoretical FP16/BF16 Tensor Core performance as a baseline [6].

### 3.3  Estimating $CO_2$ Emissions

Beyond computational time and cost, the environmental impact of training large models is a critical consideration. To quantify this, we estimate the carbon dioxide equivalent ($CO_2$e) emissions for each training process. Our approach is based on the methodology outlined by Patterson et al. [16], which combines total energy consumption with the carbon intensity of the local power grid. We calculate the total emissions in metric tons of $CO_2$e (t$CO_2$e) using the following simplified formula:

$$tCO_2e = \text{GPU-Hours} \times \text{TDP}_{(MW)} \times \text{PUE} \times \text{CI}_{\text{Grid}}$$

Where TDP is the Thermal Design Power of the GPU hardware in Megawatts, PUE is the Power Usage Effectiveness of the data center, and CIGrid is the carbon intensity of the power grid in t$CO_2$e/MWh. For our estimations, we assume a PUE of 1.1, consistent with modern hyperscale data centers, and a CIGrid of 0.429 t$CO_2$e/MWh, representing the average data center carbon emissions for 2020, as reported in the reference study [16]. This calculation allows us to translate the abstract cost of GPU-hours into a tangible measure of environmental impact, providing a more holistic view of the efficiency of different model adaptation strategies.

### 3.4 Benchmark Evaluation

To evaluate the reasoning capabilities of the models in our study, we selected two challenging benchmarks: GPQA, a graduate-level question-answering dataset [17], and AIME 2024, a competition-level math problem-solving benchmark [18]. These were chosen because they rigorously test deep reasoning and problem-solving skills, which are key capabilities that distillation aims to transfer. Furthermore, they are widely adopted in the community, with many leading proprietary and open-source models reporting scores, which provides a rich landscape for comparative analysis.

For our evaluation metric, we adopt the pass@k methodology, as it provides a more stable and robust measure of performance than greedy decoding, especially for complex reasoning tasks [19]. Specifically, we report pass@1 scores and follow the evaluation setup detailed in the DeepSeek-R1 paper [2] for consistency. We generate k responses for each question using a sampling temperature of 0.6 and a top-p value of 0.95. The pass@1 score is then calculated as the average correctness across these k samples:

$$\text{pass@1} = \frac{1}{k} \sum_{i=1}^{k} p_i$$

where $p_i$ denotes the binary correctness of the i-th response. This standardized approach ensures that our results are comparable to those published in other leading research.

### 3.5 Models and Assumptions

To ground our analysis, we center our case studies on widely recognized models and make our assumptions explicit:

- **Case Studies:** We compare vanilla (non-distilled) models directly with its distilled counterparts across benchmarks we ran locally. We also compare it with published results for larger open source and closed source models.
- **Data Precision:** All calculations assume training is performed using FP16/BF16 precision, which is standard for modern LLM training and leverages GPU Tensor Cores.
- **Efficiency Assumptions:** For our GPU-hour calculations, we use an MFU range consistent with published results, such as the 38-43% reported for Llama 3.1 on H-series GPUs [15]. For older hardware like the A-series, we assume a slightly lower MFU of 25-35%.
- **Epochs:** We assume a single epoch of training (E=1) for large-scale retraining and distillation examples, and three epochs (E=3) for common fine-tuning recipes, consistent with practices in the literature.

By clearly defining our methodology and assumptions, we provide a transparent and reproducible framework for evaluating the computational trade-offs of different model specialization strategies.

## 4    Case Studies

To demonstrate our methodology in practice, we present a series of case studies that walk through the cost estimation process for different types of models. These examples serve to ground the abstract formulas from Section 3 in concrete numbers and illustrate the fundamental differences in resource requirements between standard training and knowledge distillation.

### 4.1    The Baseline: Training a Vanilla Foundation Model

We begin by establishing a baseline using a standard, non-distilled open-source model. For this, we analyze Qwen3-8B, an 8-billion parameter model trained on a large corpus of data. This case study demonstrates the standard calculation for the full pre-training cost of a modern LLM.

**1. Training Parameters:**

- Model Size (N): 8 billion parameters
- Training Dataset (D): 36 trillion tokens

**2. FLOPs Calculation:** Using the standard formula for from Section 2.3:

$$\text{Training}_{\text{FLOPs}} = 6 \times (8 \times 10^9) \times (36 \times 10^{12}) = 1.728 \times 10^{24} \text{ FLOPs}$$

3. **GPU-Hours Calculation:** To convert FLOPs to a tangible training time, we estimate the required GPU-hours on a modern NVIDIA H100 SXM GPU. We assume an MFU of 38%, consistent with large-scale training runs.

- Peak Theoretical Performance (H100 SXM, 16-bit non-sparse): ~1,000 TFLOPs
- Model FLOPs Utilization (MFU): 38%
- Effective Performance: 1,000 TFLOPs×0.38=380 TFLOPs

$$\text{GPU−Hours} = \frac{1.728 \times 10^{24} \text{ FLOPs}}{380 \times 10^{12} \text{FLOPs/s} \times 3600 \text{ s/hr}} \approx 1{,}263{,}158 \text{ hours}$$

**4. $CO_2$ Emissions Calculation:** Finally, we estimate the environmental impact using the GPU's Thermal Design Power (TDP) of 700W (0.0007 MW).

- Total Energy: 1,263,158 h $\times$ 0.0007 MW $\times$ 1.1 PUE $\approx$ 972.6 MWh
- Total Emissions: 972.6 MWh $\times$ 0.429 $tCO_2$e/MWh $\approx$ 417.25 $tCO_2$e

This baseline demonstrates that training even a relatively small 8B parameter model from scratch requires a staggering investment of over 1.2 million H100 GPU-hours and results in over 417 metric tons of $CO_2$ equivalent emissions.

### 4.2    The Challenger: Creating a Distilled Model

To contrast with the baseline, we now analyze the process of creating a distilled model of the same size: the 8B parameter DeepSeek-R1-0528-Qwen3-8B. This model inherits its capabilities from a much larger, 685B parameter "teacher" model. This case study highlights the dramatically different cost structure of distillation.

**1. Distillation Parameters:**

- Teacher Model Size ($N_{teacher}$): 685 billion parameters
- Student Model Size ($N_{student}$): 8 billion parameters
- Distillation Dataset ($D_{distill}$): 800,000 samples, which we estimate ~600M tokens.

**2. FLOPs Calculation:** We use the two-part distillation formula from Section 2.3:

- Teacher Inference: $2 \times (685 \times 10^9) \times (600 \times 10^6) = 8.22 \times 10^{20}$ FLOPs
- Student Training: $6 \times (8 \times 10^9) \times (600 \times 10^6) = 2.88 \times 10^{19}$ FLOPs
- Total Distillation FLOPs: $(8.22 \times 10^{20}) + (2.88 \times 10^{19}) \approx 8.5 \times 10^{20}$ FLOPs

**3. GPU-Hours Calculation:** Using the same H100 GPU and MFU as the baseline:

$$\text{GPU−Hours} = \frac{8.5 \times 10^{20} \text{ FLOPs}}{380 \times 10^{12} \text{FLOPs/s} \times 3600 \text{ s/hr}} \approx 621.9 \text{ hours}$$

**4. $CO_2$ Emissions Calculation:**

- Total Energy: 621.9 h $\times$ 0.0007 MW $\times$ 1.1 PUE $\approx$ 0.479 MWh
- Total Emissions: 0.479 MWh $\times$ 0.429 $tCO_2e$/MWh $\approx$ 0.205 $tCO_2e$

The comparison is stark. Creating a highly capable 8B parameter model via distillation requires approximately 622 H100 GPU-hours, a reduction of over 2,000 times compared to training its vanilla counterpart from scratch. The corresponding carbon footprint is also reduced from over 400 metric tons to less than a single ton, illustrating the profound efficiency gains offered by this approach.

### 4.3    The Specialization Alternative: Fine-Tuning

We also analyze a third common scenario: specializing a pre-trained model for a particular domain via fine-tuning. This represents a highly compute-efficient path to task-specific performance, but with a different set of trade-offs compared to distillation.

**1. Fine-Tuning Parameters:**

- Model Size (N): 8 billion parameters
- Domain-Specific Dataset (D): 10 billion tokens

- Epochs (E): 3

**2. FLOPs Calculation:** Using the standard training formula, we account for the multiple epochs common in fine-tuning recipes.

$$\text{Training}_{\text{FLOPs}} = 6 \times (8 \times 10^9) \times (10 \times 10^9) \times 3 = 1.44 \times 10^{21} \text{ FLOPs}$$

**3. GPU-Hours Calculation:** Using the same H100 GPU and MFU as the previous examples:

$$\text{GPU-Hours} = \frac{1.44 \times 10^{21} \text{ FLOPs}}{380 \times 10^{12} \text{ FLOPs/s} \times 3600 \text{ s/hr}} \approx 1052.6 \text{ hours}$$

**4. $CO_2$ Emissions Calculation:**

- Total Energy: 1,052.6 h $\times$ 0.0007 MW $\times$ 1.1 PUE $\approx$ 0.81 MWh
- Total Emissions: 0.81 MWh $\times$ 0.429 t$CO_2$e/MWh $\approx$ 0.347 t$CO_2$e

From a pure resource perspective, fine-tuning is also exceptionally efficient, requiring a fraction of the compute of full pre-training. However, its scope is inherently limited. This process adapts an existing model to a narrow domain, but it cannot impart the broad, generalizable reasoning capabilities that distillation can transfer from a much larger and more powerful teacher. Therefore, while fine-tuning is optimal for narrow task adaptation, distillation provides a superior path for creating smaller models that retain a high degree of general intelligence.

### 4.4  Estimating the Proprietary Frontier

To complete our comparative landscape, we analyze a state-of-the-art proprietary model, GPT-4o (0513 version). While exact architectural details remain undisclosed, external analyses provide credible estimates for the total training compute [20]. This allows us to apply our methodology to approximate the resources required to create a frontier model.

**1. Training Parameters:** Total Estimated Compute: $3.8 \times 10^{25}$ FLOPs [20]

**2. GPU-Hours Calculation:** As the model size and training data are unknown, we directly use the estimated total FLOPs. Applying the same H100 hardware profile as in previous cases:

$$\text{GPU-Hours} = \frac{3.8 \times 10^{25} \text{ FLOPs}}{380 \times 10^{12} \text{ FLOPs/s} \times 3600 \text{ s/hr}} \approx 27{,}777{,}778 \text{ hours}$$

**3. $CO_2$ Emissions Calculation:**

- Total Energy: 27,777,778 h × 0.0007 MW × 1.1 PUE ≈ 21,388.9 MWh
- Total Emissions: 21,388.9 MWh × 0.429 tCO$_2$e/MWh ≈ 9,175.83 tCO$_2$e

The scale of these figures is immense. The estimated cost to train a model like GPT-4o is over 27 million H100 GPU-hours, producing over 9,000 metric tons of CO$_2$e. This is more than 20 times the cost of training the 8B parameter Qwen3 model and underscores the astronomical resource barrier to entry for developing frontier models. It provides a powerful context for our results, reinforcing the critical importance of compute-efficient methods like distillation for enabling broader access to high-performance AI.

## 5    Results

This section presents the aggregated results of our analysis, consolidating the estimated costs and benchmark performance across a wide range of models. We first provide a comprehensive comparison of all models in our study, followed by a graphical analysis of performance versus compute, and conclude with a practical look at how hardware choice impacts training time.

### 5.1    Comprehensive Model Comparison

Table 1 provides a detailed breakdown of the models analyzed in this study. The performance data presented is a composite of our own local evaluations and publicly reported scores. For the first six models listed, we ran benchmarks locally using established evaluation frameworks such as LocalAIME [21], OpenAI's Simple Evals [22], and community-developed scripts [23]. For all other models, performance data was collected from official technical reports, model cards, leaderboards, and other reputable public sources [24-30].

**Table 1.** Comprehensive model comparison and benchmarks evaluation

| Model | Type | Training Compute | GPU-Hours (H100) | tCO$_2$e | GPQA Diamond | AIME 2024 |
|---|---|---|---|---|---|---|
| Llama 3.1 8B | Vanilla | 7.49E+23 | 1.46E+06 | 4.20E+02 | 30.4 | 10 |
| DeepSeek-R1-Distill-Llama-8B | Distilled | 8.34E+20 | 6.10E+02 | 2.01E-01 | 49 | 50.4 |
| Llama 3.3 70B Instruct | Vanilla | 6.55E+24 | 7.00E+06 | 2.04E+03 | 46.7 | 22.5 |
| DeepSeek-R1-Distill-Llama-70B | Distilled | 1.06E+21 | 7.73E+02 | 2.55E-01 | 65.2 | 70 |
| Qwen 3 8B | Vanilla | 1.73E+24 | 1.26E+06 | 4.17E+02 | 62 | 76 |
| DeepSeek-R1-0528-Qwen3-8B | Distilled | 8.51E+20 | 6.22E+02 | 2.05E-01 | 61.1 | 86 |
| Phi-4 14B | Vanilla | 9.30E+23 | 6.80E+05 | 2.25E+02 | 69.3 | 81.3 |
| GPT-4o 0513 | Proprietary | 3.80E+25 | 2.78E+07 | 9.18E+03 | 53.6 | 9.3 |
| Deepseek V3 | Vanilla | 3.40E+24 | 2.49E+06 | 8.21E+02 | 59.1 | 39.2 |
| Grok 3 | Proprietary | 4.60E+26 | 3.36E+08 | 1.11E+05 | 75.4 | 52.2 |
| Claude 3.7 Sonnet | Proprietary | 3.30E+25 | 2.41E+07 | 7.97E+03 | 68 | 23.3 |
| o3-mini | Proprietary | 4.68E+23 | 3.80E+05 | 1.26E+02 | 76.8 | 79.6 |
| Qwen3-235B-A22B | Vanilla | 5.08E+25 | 3.71E+07 | 1.23E+04 | 71.1 | 85.7 |
| Llama 3.1 405B | Vanilla | 3.79E+25 | 3.08E+07 | 8.93E+03 | 50.7 | 23.3 |
| Deepseek-R1 | Teacher | 5.20E+24 | 3.80E+06 | 1.26E+03 | 71.5 | 79.8 |
| Deepseek-R1-0528 | Teacher | 5.20E+24 | 3.80E+06 | 1.26E+03 | 81 | 91.4 |

For each model, we list its type—distinguished here as "Vanilla" for open-source models trained from scratch, "Distilled" for models created via knowledge distillation, and "Proprietary" for closed-source models. The table includes the total training compute in FLOPs, our estimated training time in H100 GPU-hours, the corresponding $CO_2$ emissions, and the model's performance on the GPQA and AIME 2024 benchmarks. Several key comparisons emerge directly from this data. For instance, the vanilla Llama 3.1 8B model required approximately 1.46 million H100 GPU-hours to train, whereas its distilled counterpart, DeepSeek-R1-Distill-Llama-8B, achieved superior performance on both benchmarks with only ~610 H100 GPU-hours—a reduction in compute of over 99.9%. This pattern of multi-order-of-magnitude efficiency gains for distilled models is consistent across all comparable pairs in our analysis.

### 5.2   Performance vs. Compute Analysis

To visualize the relationship between computational investment and reasoning performance, we plot the accuracy of each model against its total training compute. Fig 2 and Fig 3 show the results for the GPQA and AIME benchmarks, respectively, with the training compute axis on a logarithmic scale to accommodate the wide range of values.

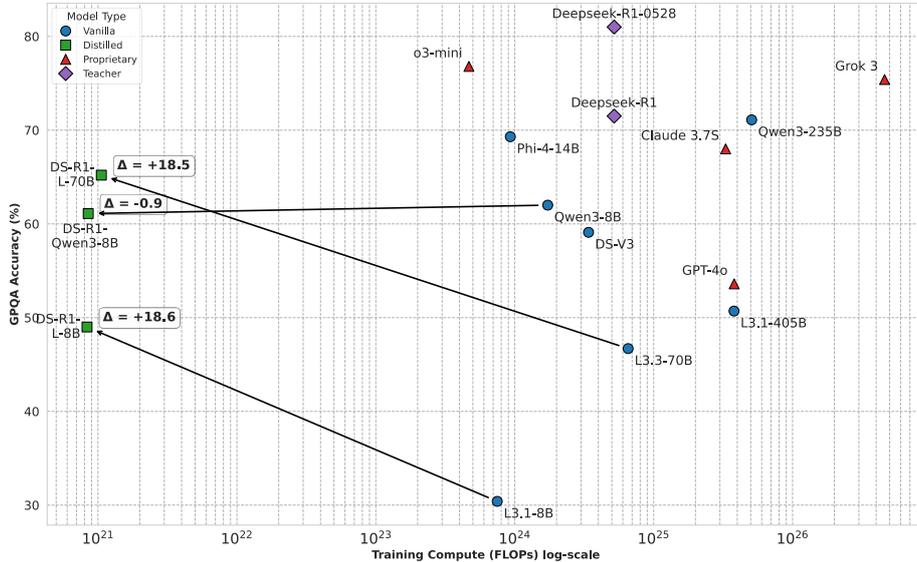

**Fig. 2.** GPQA accuracy v Training Compute (FLOPs) log-scale for models

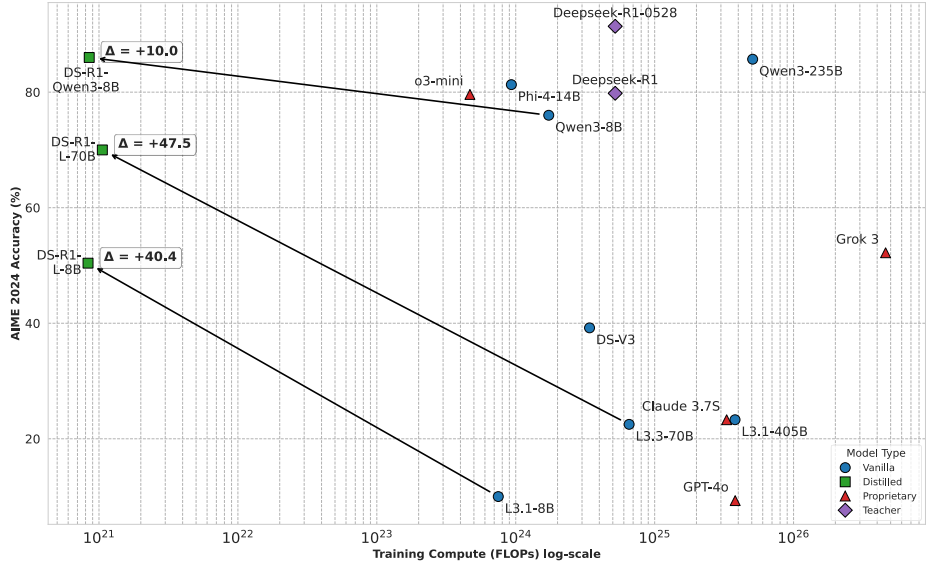

**Fig. 3.** AIME 24 accuracy v Training Compute (FLOPs) log-scale for models

The plots clearly reveal several key trends, with the AIME 2024 results in Figure 2 providing a particularly stark illustration. While there is a general positive correlation between training compute and benchmark performance for vanilla models, the distilled models establish a new, far more efficient performance curve. For example, the distilled DeepSeek-R1-0528-Qwen3-8B model achieves a remarkable 86% accuracy. This not only dramatically exceeds its vanilla 8B counterpart but also surpasses the performance of the massive Qwen3-235B model (85.7%), which required nearly 60,000 times more training compute to develop. This trend demonstrates that for complex reasoning tasks, distillation is not merely an optimization but a method for achieving a step-change in performance per FLOP, challenging the notion that frontier performance is exclusively tied to frontier-scale compute.

### 5.3  Hardware Impact on Training Time

**Table 2.** Comparison of performance and $CO_2$e across GPUs

| Workload | Training Compute | A100 (80GB) | L40S (48GB) | H100 SXM (80GB) |
|---|---|---|---|---|
| | | Hours \| $tCO_2$e | Hours \| $tCO_2$e | Hours \| $tCO_2$e |
| Qwen3-8B Training (Vanilla) | 1.73E+24 | 4.67M \| 881.3 | 4.42M \| 730.0 | 1.26M \| 417.3 |
| 8B Model Distillation (Challenger) | 8.50E+20 | 2297 \| 0.43 | 2174 \| 0.36 | 622 \| 0.21 |
| 8B Model Fine-Tuning (Specialization) | 1.44E+21 | 3890 \| 0.73 | 3683 \| 0.61 | 1053 \| 0.35 |

While H100 GPUs were used as the baseline for our primary calculations, the choice of hardware has a significant practical impact on project timelines and costs. Table 2 provides a focused comparison of estimated training times and $CO_2$ footprints for a selection of models across different generations of NVIDIA data center GPUs. This table illustrates that for a fixed computational workload (FLOPs), newer hardware like the H200 can reduce training time and associated emissions compared to previous generation A100 GPUs, highlighting the role of hardware efficiency in the overall cost equation.

The collective results from our tables and plots provide strong, quantitative evidence that knowledge distillation represents a paradigm shift in creating compute-efficient, high-performance language models.

## 6  Discussion

Our analysis of the computational costs and performance benchmarks associated with various model creation strategies provides a clear, quantitative basis for decision-making. The results from Section 5, particularly the comprehensive data in Table 1 and the performance-compute relationships visualized in Figures 2 and 3, surface critical insights into the trade-offs between these methods.

### 6.1  Deconstructing the Cost-Performance Trade-off

The case studies in Section 4 established the vast differences in computational cost between model creation methods, a finding that is reinforced across the entire landscape of models in Table 1. Training a vanilla foundation model from scratch, even a relatively small 8B parameter model, requires over a million GPU-hours and hundreds of tons of $CO_2$e. In stark contrast, creating a distilled model of the same size, as shown in our case study, requires less than 0.1% of that compute budget.

However, the most crucial finding of this paper is not just that distillation is cheaper, but that it redefines the efficiency frontier. As visualized in Figures 2 and 3, distilled models do not simply follow the same scaling curve as vanilla models. Instead, they establish a new, superior curve. A distilled 8B model, for example, can achieve reasoning performance on par with, or even exceeding, a vanilla 70B model, but with a computational cost that is orders of magnitude lower. This demonstrates that distillation is not merely a cost-saving measure; it is a method for achieving a step-change in performance per FLOP. Fine-tuning, while also computationally inexpensive, does not exhibit this trait, as its performance is fundamentally capped by the capabilities of its base model. Distillation breaks this ceiling by directly transferring knowledge from a far more capable teacher.

### 6.2  The Strategic Imperative for Distillation

Based on these results, the strategic context for employing distillation becomes clear. It is the optimal choice when the goal is to create a deployable model that maximizes reasoning capability within a constrained computational budget. The key prerequisites are the existence of a powerful teacher model—whether proprietary or a large internal model—and a well-curated distillation dataset.

Our findings suggest that for tasks demanding high-level reasoning, investing a small fraction of a compute budget into a distillation process can yield a significantly higher return on investment than spending the same budget on further pre-training or fine-tuning a vanilla model. This holds true across the model size spectrum, from creating small, on-device models to developing highly capable mid-size models for enterprise use. Furthermore, as noted in our methodology, the high initial cost of the teacher's inference pass is a one-time, amortizable expense. Organizations can reuse the resulting dataset to distill multiple specialized student models, further driving down the effective cost per model and enabling the creation of a diverse suite of efficient, high-performing AI agents.

### 6.3   Broader Implications for the AI Ecosystem

The practical implications of these findings are significant. The extreme resource requirements for training frontier models, as estimated for GPT-4o in Section 4.4, create a landscape dominated by a few large players. Knowledge distillation is a powerful democratizing force. It provides a viable pathway for startups, academic institutions, and enterprises without access to massive GPU clusters to create models with competitive, state-of-the-art reasoning abilities.

This has profound effects on several fronts:

- **Economic Accessibility:** By drastically lowering the barrier to entry for creating high-performance models, distillation fosters innovation and competition. The estimated 1.2M GPU-hours for training a vanilla 8B model represents a significant financial and environmental cost, whereas the ~622 GPU-hours for its distilled counterpart are orders of magnitude more accessible.
- **Deployment Flexibility:** The smaller footprint of distilled models makes them ideal for a wider range of deployment scenarios, from latency-sensitive cloud applications to privacy-centric on-premise or edge deployments.
- **Sustainable AI Development:** As the environmental cost of AI becomes a greater concern, methods that dramatically reduce the carbon footprint of model creation are critical. As our results show, the difference in emissions between training from scratch and distilling is substantial, making distillation a key strategy for more sustainable AI practices.

In essence, our results quantitatively demonstrate that distillation is not just an optimization technique but a strategic necessity for any organization looking to balance performance, cost, and accessibility in the development of advanced AI systems.

## 7  Conclusion

This paper set out to benchmark the performance and efficiency of distilled language models in resource-constrained settings. Our comprehensive analysis, spanning computational cost estimation, local benchmarking, and synthesis of public data, confirms that knowledge distillation offers a transformative approach to model creation. The findings presented in our results demonstrate that distilled models consistently outperform their vanilla counterparts, achieving superior reasoning capabilities for a fraction of the training cost.

For practitioners operating under computational constraints, this trade-off is paramount. The dual benefits of distillation—drastically reduced training compute and the creation of smaller, more performant models—directly address the primary challenges of real-world deployment. Crucially, the efficiency gains extend beyond the training phase. Since inference latency is fundamentally linked to the size of a model and its active parameters, the smaller student models produced through distillation are inherently faster and cheaper to run. This enables the deployment of powerful AI in latency-sensitive applications where larger models would be impractical. Ultimately, by providing a pathway to elite performance without demanding elite resources, distillation is the definitive strategy for developing and deploying advanced, efficient, and accessible language models.